\documentclass[journal]{IEEEtran}
\usepackage{comment}
\usepackage{graphicx}
\usepackage{adjustbox}
\usepackage{multirow}
\usepackage{ctable}
\usepackage[switch]{lineno}

\newcommand{\cmmnt}[1]{\ignorespaces}

\ifCLASSINFOpdf
\else
\fi
\begin{document}
%
\title{Machine Learning-based Lie Detector applied to a Novel Annotated Game Dataset}
%
%
%

\author{Nuria~Rodriguez Diaz,~
        Decky~Aspandi,~
        Federico~Sukno,
        and~Xavier~Binefa
\thanks{Nuria~Rodriguez~Diaz, D.\ Aspandi, F.\ Sukno and X.\ Binefa are with the Department
of Information and Communication Technology, Universitat Pompeu Fabra, Barcelona, Spain, 08026. 
E-mail: nuriarodriguezdiaz@gmail.com}
\thanks{D.\ Aspandi is also with Institute for Parallel and Distributed Systems,  University of Stuttgart, Stuttgart, Germany, 70569. E-mail: decky.aspandi-latif@ipvs.uni-stuttgart.de
}
}

%
%

\markboth{Journal of \LaTeX\ Class Files,~Vol.~14, No.~8, August~2015}%
{Shell \MakeLowercase{\textit{et al.}}: Bare Demo of IEEEtran.cls for IEEE Journals}
%



\maketitle


\begin{abstract}
Lie detection is considered a concern for everyone in their day to day life given its impact on human interactions. Thus, people normally pay attention to both what their interlocutors are saying and also to their visual appearances, including faces, to try to find any signs that indicate whether the person is telling the truth or not. While automatic lie detection may help us to understand this lying characteristics, current systems are still fairly limited, partly due to lack of adequate datasets to evaluate their performance in realistic scenarios. In this work, we have collected an annotated dataset of facial images, comprising both 2D and 3D information of several participants during a card game that encourages players to lie. Using our collected dataset, We evaluated several types of machine learning-based lie detectors in terms of their generalization, person-specific and cross-domain experiments. Our results show that models based on deep learning achieve the best accuracy, reaching up to 57\% for the generalization task and 63\% when dealing with a single participant. Finally, we also highlight the limitation of the deep learning based lie detector when dealing with cross-domain lie detection tasks. 


\end{abstract}


%
\IEEEpeerreviewmaketitle

\section{Introduction}
%
%
%
%

\IEEEPARstart{I}{t} is very hard for humans to detect when someone is lying. Ekman \cite{ekman_fme}. highlights five reasons to explain why it is so difficult for us: $1)$ the fact that during most human history, there were smaller societies in which liars would
have had more chances of being caught with worse consequences than nowadays, $2)$
that children are not taught on how to detect lies, since even
their parents want to hide some things from them; $3)$ that people prefer to trust in what they are told; $4)$ or prefer not to know the real truth; and $5)$ that people are taught to be polite and \emph{not steal information that is not given to us}. However, it has been argued that it is possible for someone to learn how to detect lies in another person given sufficient feedback (e.g. that 50$\%$ of the times that person is lying) and focusing on micro-expressions \cite{ekman_fme,haggard_fme} 

Building from the above, the detection of deceptive behavior using facial analysis has been proved feasible using macro- and, especially, micro-expressions \cite{ddvideos,multimodaldd,face-focused}. However, micro-expressions are difficult to capture at standard frame rates and, given that humans can learn how to spot them to perform lie detection, the same training might be used by liars to learn how to hide them. Thus, there has been interest in detecting facial patterns of deceptive behavior that might not be visible to the naked eye, such as the heat signature of the periorbital- \cite{tsiamyrtzis2007imaging} or perinasal-region \cite{dcosta2015perinasal} in thermal imagery, which cannot be perceived by human vision.

One of the crucial aspects to appropriately address lie-detection research is the availability of adequate datasets. The acquisition of training and, especially, evaluation material for lie detection is a challenging task, particularly regarding the necessity to gather ground truth, namely, to know whether a person is lying or not. The main difficulty arises because such knowledge is not useful if the scenario is naively simulated (e.g. it is not sufficient to instruct a person to simply tell a lie). Research on high-stakes lies suggests that deceptive behavior can depend heavily on the potential consequences for the liar \cite{porter2010truth}. Thus, researchers have attempted to create artificial setups that can convincingly reproduce situations where two factors converge: 1) there is a potential for truthful deceptive behavior; 2) we know when deception takes place and when the recorded subjects are telling the truth. Most attempts so far have focused on interview scenarios in which the participants are \emph{instructed to lie} \cite{dcosta2015perinasal,tsiamyrtzis2007imaging,mohamed2006brain}, although it is hard to simulate a realistic setting for genuine deceptive behavior. Alternatively, some researchers have worked in collaboration with Police Departments, with the benefit of a scenario that in many cases is 100$\%$ realistic, as it is based on interviews to criminal suspects. However, the problem in this setting is the ground truth: it is not possible to rely on legal decision-making \cite{vrij2008detecting} and even the validity of confessions has been questioned \cite{frank2008human}.

In contrast, in this paper we explore an alternative scenario in which participants are recorded while playing a competitive game in which convincingly lying to the opponent(s) produces an advantage. On one hand, participants are intrinsically motivated to lie convincingly. And, importantly, given the knowledge of the game rules, we can accurately determine whether a given behavior is honest or deceiving. The use of card games can also benefit from the occurrence of unexpected events that produce genuine surprise situations to the potential liar, which and has been highlighted as beneficial for lie detection scenarios \cite{porter2010truth}. 

Thus, the goals of this paper are two-fold. Firstly, we present an annotated dataset, the Game Lie Dataset (GLD), based on frontal facial recordings of 19 participants who try their best to fool their opponents in the \emph{liar} card game. Secondly, we depart from the dominating trend of lie-detection based on micro-expressions and investigate whether a lie can be detected by analyzing solely the facial patterns contained on single images as input to cutting edge machine learning \cite{microexpressions1,action_units_rgb,pain_expressions} and deep learning \cite{aspandi2021enhanced,comas2020end,aspandi2020latent,aspandi2019fully} facial analysis algorithms.

Using our collected dataset and several automatic lie detection models, we perform lie detection experiments under 3 different settings: $1)$ Generalization test, to evaluate the performance on unseen subjects; $2)$ Person specific test, to evaluate the possibility to learn how a given participant would lie, and $3)$ Cross-domain test, to evaluate how the models generalize to a different acquisition setup. The contributions of this work can be summarized as follows:
\begin{enumerate}
    \item We present the GLD dataset, which contains coloured facial data as well as ground truth (lie/true) annotations, captured during a competitive card game in which participants are rewarded for their ability to lie convincingly. 
    \item We also present quantitative comparisons results of several machine learning models tested on the new captured dataset. 
    \item We present several experiments that outline the current limitations of facial-based lie detection when dealing with several different lie tasks. 
\end{enumerate}


\section{Related work}

Different approaches and techniques have been applied for the lie detection task, with Physiological cues widely and commonly used. The most popular one is the polygraph, commonly known as a lie detection machine. Other approaches have used brain activity in order to detect deception by utilising different neuro-imaging methods such as fMRI \cite{fmri1,fmri2,memory,mohamed2006brain}. For example, Markowitsch \cite{memory} compared brain scans from volunteers in a lie-detection experiment, in which some participants were asked to lie and others had to tell the truth. It was found that when people were telling the truth, the brain region associated with sureness was activated, while in the case of lies the area associated with mental imagination was activated. Similarly, brain's hemoglobin signals (fNIRS) or electrical activity (EEG) can be measured to define physiological features for lie detection (\cite{fnirs1}, \cite{fnirs2}, \cite{fnirs3}, \cite{eeg}).

The main drawback of the above techniques, however, is their invasive and expensive nature due the need for special instruments to allow the data collections. This has led to the emergence of less invasive approaches involving verbal and non-verbal cues. Several studies focus on utilising thermal imaging to perform the deception detection task, since skin temperature has been shown to significantly rise when subjects are lying (\cite{thermal_airport,perinasal}). Furthermore, the speech has also been explored (\cite{multilingual,multimodal_truth}, e.g. by extracting features based on transcripts, part of speech (PoS) tags, or acoustic analysis (Mel-frequency cepstral coefficients).

The use of several modalities for lie detection has also been investigated to see its impact in improving the detection algorithms.  In \cite{verbal,real-life_trial,boxoflies} both verbal and non-verbal features were utilised. The verbal features were extracted from linguistic features in transcriptions, while the non-verbal ones consisted in binary features containing information about facial and hands gestures. In addition, \cite{boxoflies} introduced dialogue features, consisting in interaction cues. Other multi-modal approaches have combined the previously mention verbal and non-verbal features together with micro-expressions (\cite{ddvideos}, \cite{multimodaldd}, \cite{face-focused}), thermal imaging ( \cite{multimodaldd2} ), or spatio-temporal features extracted from 3D-CNNs (\cite{multimodal_deep}, \cite{multimodal_deep2}).

In the last decade, there has been a growing interest in the use of facial images to perform lie detection, often based on micro-expressions \cite{microexpressions1,pain_expressions,ddvideos,multimodaldd,face-focused} or facial action units \cite{action_units_rgb}, achieving the current state of the art accuracy.

\subsection{Existing Lie detection datasets}

Despite the existing works to perform lie detection tasks, just a few datasets are published. In the literature, there are only two existing multi-modal, audio-visual datasets that are specifically constructed for the purpose of lie detection tasks: A multi-modal dataset based on the Box-of-Lies\textregistered \ TV game (\cite{boxoflies}) and a multi-modal dataset using real life Trial-Data (\cite{real-life_trial}).

Both the Box-of-Lies and Trial-Data include 40 labels for each gesture a participant shows and the whole transcripts for all videos. The difference between them lies in the interactions: in the Trial data there is only a single speaker per video and lies are judged from the information of this single speaker. In contrast, in the Box-of-Lies\textregistered,  the lies are identified from the interaction between two people while playing a game, with emphasis on their dialogue context. Thus, the Box of Lies\textregistered \ dataset also contains annotations on participants feedback, in addition to veracity tags for each statement made.

Even though previous datasets have provided a way to analyse the respective lying characteristics, there still exist some limitations: the first one is that the interactions between participants are fairly limited, which are usually constrained to one to one lying setting. Furthermore, the faces are usually taken on extremely different settings and pose which may hinder the model learning. In this work, we present a novel dataset that involves more interactions between participants during the lying. We also record our data on a controlled settings (environment) to reduce invariability of irrelevant image characteristics such as lighting and extreme poses to allow for more precise machine learning based modelling.



\begin{table*}[t]
\caption{Existing Lie detection dataset.}
\resizebox{\textwidth}{!}{\begin{tabular}{l|cccc|ccccc|c}
\toprule
\multirow{2}{*}{\textbf{Dataset}} & \multicolumn{4}{c|}{\textbf{Subjects}}      & \multicolumn{5}{c|}{\textbf{Videos}}& \multirow{2}{*}{\textbf{Year}} \\
\cline{2-10} 
& \textbf{Total} & \textbf{M} & \textbf{F} & \textbf{Age range} & \textbf{Total} & \textbf{Utterances} & \textbf{Deceptive} & \textbf{Truthful} & \textbf{Duration} & \\ \hline
Box-of-Lies \cite{boxoflies}& 26& 6 & 20& No Information        & 25    & 1049       & 862       & 187      & 144 minutes       & 2019\\ 
Trial-Data \cite{real-life_trial} & 56    & 35& 21& 16-60     & 121   & 121        & 61        & 60       & None       & 2016\\
\bottomrule
\end{tabular}}
\end{table*}

\section{Deception Dataset}
In order to establish an appropriate scenario to perform the lie actions, we opt to use a card game called "The Liar" due to the unique characteristics of this game that incentivise the participants to lie well in order to win the game. Furthermore, its simplicity and easy to learn aspect allow for more efficient data collection. The winner of this game is the first participant to run out of cards.

Specifically, the game consists in dealing all cards among three or more players. 
In theory, players must throw as many cards as they want as long as all of them have the same number. However, cards are turned face down and thus, players can lie on the number in the cards.
The game round starts when a player throws some cards 
and then, the player on the right decides whether to believe the previous player or not. 
If the next player believes the previous player, he/she has to throw some cards stating that they have the same number as the ones already thrown.
If, on the contrary, the next player does not believe the previous player, the thrown cards are checked. Finally, if the previous player was telling the truth the current player has to take the cards, otherwise, the previous player will take the cards back. Thus all players are encouraged to perform the lies well in order to quickly reduce as many cards as possible.

\begin{figure}[t]
\centering
\includegraphics[width=0.45\textwidth]{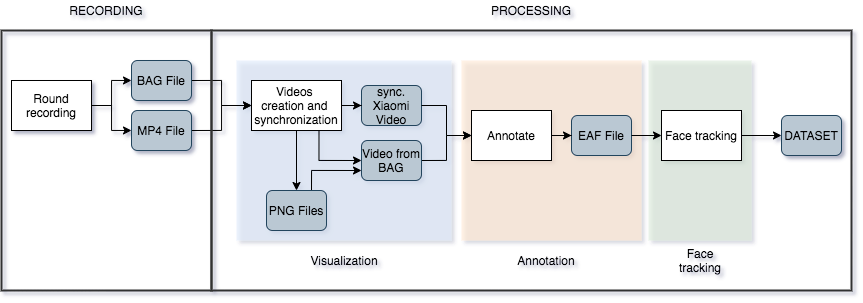}
\caption{Data processing workflow diagram.}
\label{fig:processing}
\end{figure}

These interactions between several players, along with the incentive to lie, enable us to observe the certain gestures that people exhibit in performing the lies. Furthermore, the interactions between players also allow us to include the dynamic as time progress. The general workflow used to record this game is shown in the Figure \ref{fig:processing} that we will explain in the following sections.

\subsection{Materials}

\begin{figure}[t]
\begin{center}
\includegraphics[width=0.45\textwidth]{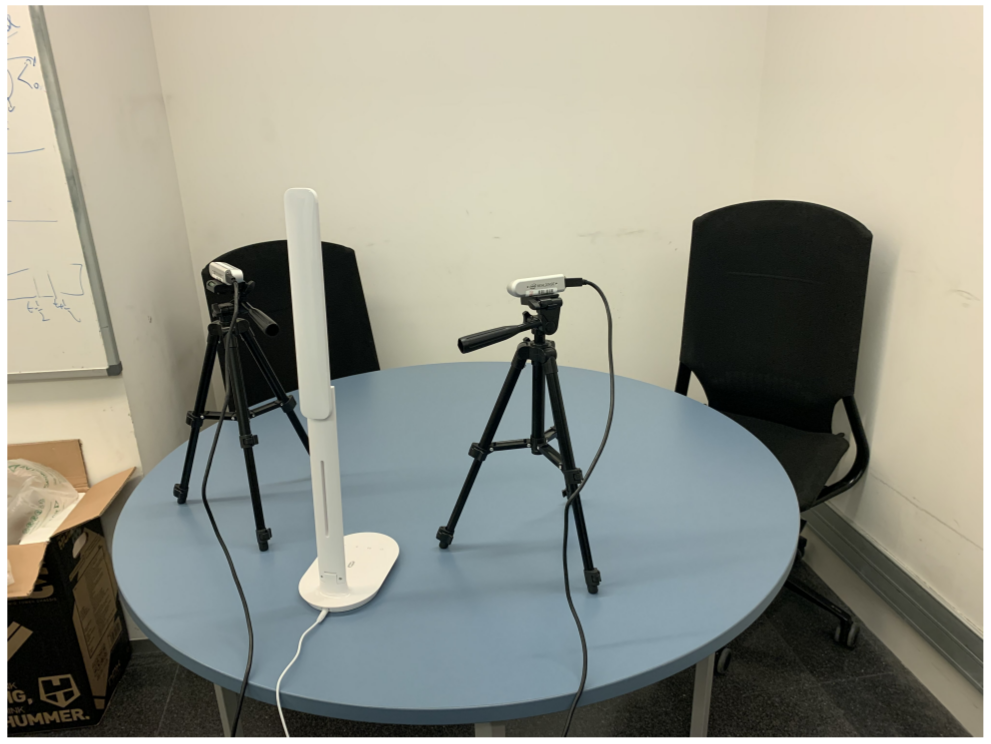}
\caption{The example of the place setup for data acquisition.}
\label{fig:setup}
\end{center}
\end{figure}

We use these materials to perform the data collections: a deck of cards for the game scenario, an RGB color camera for faces recording, a video camera for cards recording and a pair of lamps to improve the light conditions. Specifically, we operate two Intel RealSense Cameras D415 For faces recording with a frame rate of 30 fps for RGB images. For game cards recording, two video cameras Mi Action Camera 4K by Xiaomi were used. The overall table setup for the data recording can be seen in Figure~\ref{fig:setup}

\subsection{Participants}
We recorded a total of 19 participants, 8 male and 11 female. The participants are mixed graduate and undergraduate from different universities and from diverse study areas (background). The age range of the participants is between 21 and 26 years old, and they are expressing themselves in Spanish and Catalan throughout the data collections and interactions. Lastly, we have explicit consents from the all participants to use and analyse the recorded facial images for research purpose.

\subsection{Data Collection}
We perform the data collection in a total of eight sessions including number of participants that are assigned to the different groups. These groups varied between 3 and 6 participants and several rounds of game playing were performed in every session. Furthermore, two participants were recorded at a time in each round. The scenario is set such that each camera was able to record a single face from the front and other video cameras are located next to the recorded players' hands, on order to record their cards. This allows us to listen to the players statements and determine if they are lying according to the cards in the recording, which is crucial during the annotation process.

\subsection{Data Annotation and Pre-processing}

We begin our data annotation and pre-processing task by synchronizing our recorded videos of face and corresponding cards. This is done in order to determine if the corresponding player is lying. These synchronized videos are subsequently annotated with ELAN software to create comment stamps in a selected space of time. Together with these annotations, we are able to find the statements' that are in correspondent with proper frames. Finally, we extract the facial area using \cite{kazemi2014one} using relevant RGB frames and cropped the to be saved as an image in the final collected dataset, as well as a point-cloud file. 

\subsection{Dataset Contents}

\begin{figure}
    \begin{center}
    \includegraphics[height=2.5in]{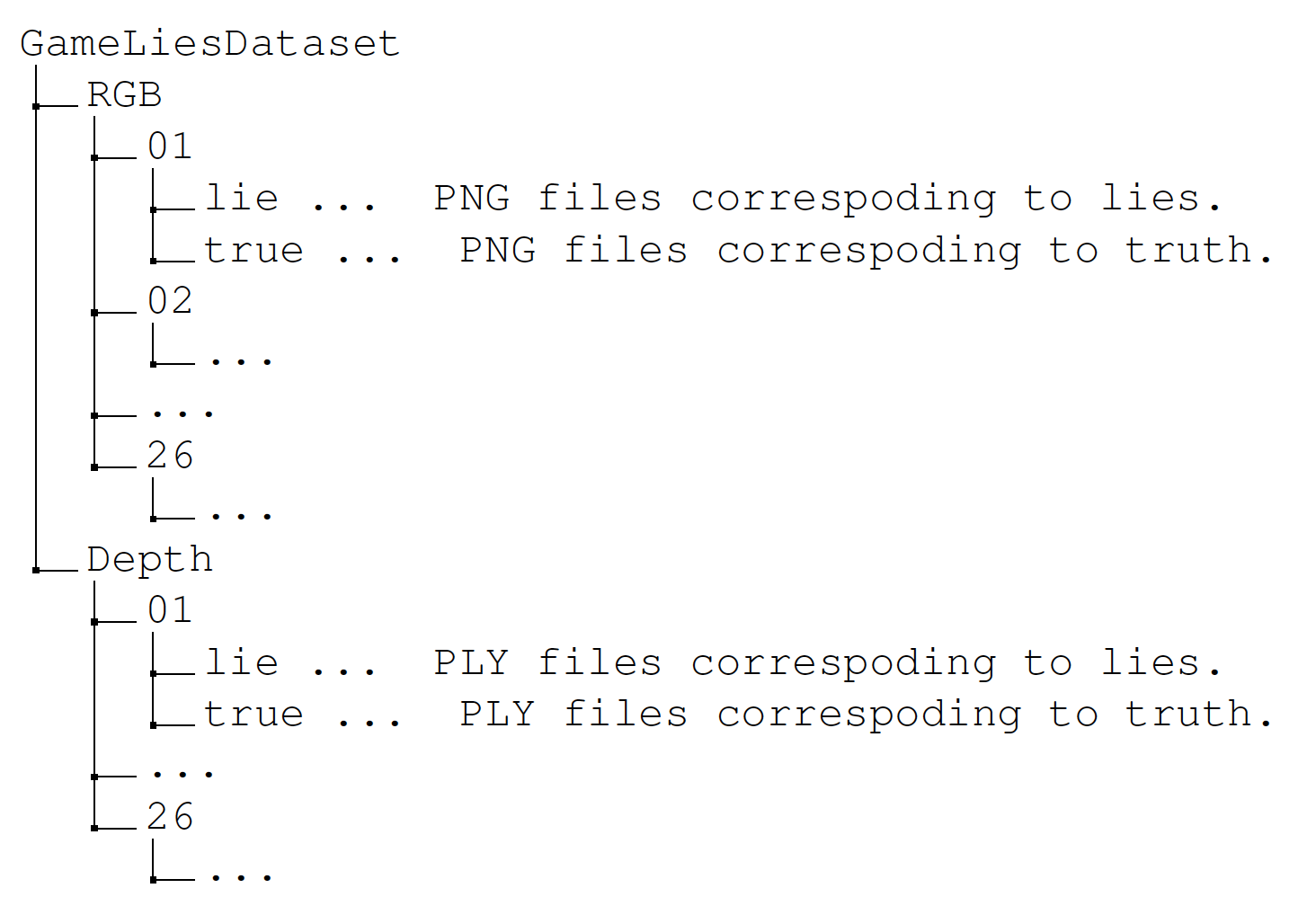}
    \caption{The structured folder of our collected dataset.}
    \label{fig:folders}
    \end{center}
\end{figure}

We create a structured folder (as seen in Figure \ref{fig:folders}) to ease future data loading and understanding during dissemination, with all recorded data stored to a root folder named \textit{Game Lies Dataset}. Both images and 3D objects are named following a convention as follow: 1\_2.PNG or 3\_4.PLY. Where the first number (1 and 3 in the example) correspond to the number of the statement and the second number (2 and 4) is the corresponding statement frame. In this instance, the PNG example corresponds to the second frame of the first statement made by the participant in the recording.

In the end, our collected \textit{Game Lies Dataset} or \textbf{GLD} contains data from 26 recordings with 18 different faces and a total number of frames of 15566, 6476 of which correspond to lies (41.6\%) and 9090 to true (58.4\%). This frames correspond to a total of 417 statements, 170 of which are lies (40.8\%) and 247 are true (59.2\%). Hence on average, each lie statement has 38 frames, and true statements consisting about 37 frames.

\begin{figure}[t]
    \begin{center}
    \includegraphics[height=1in]{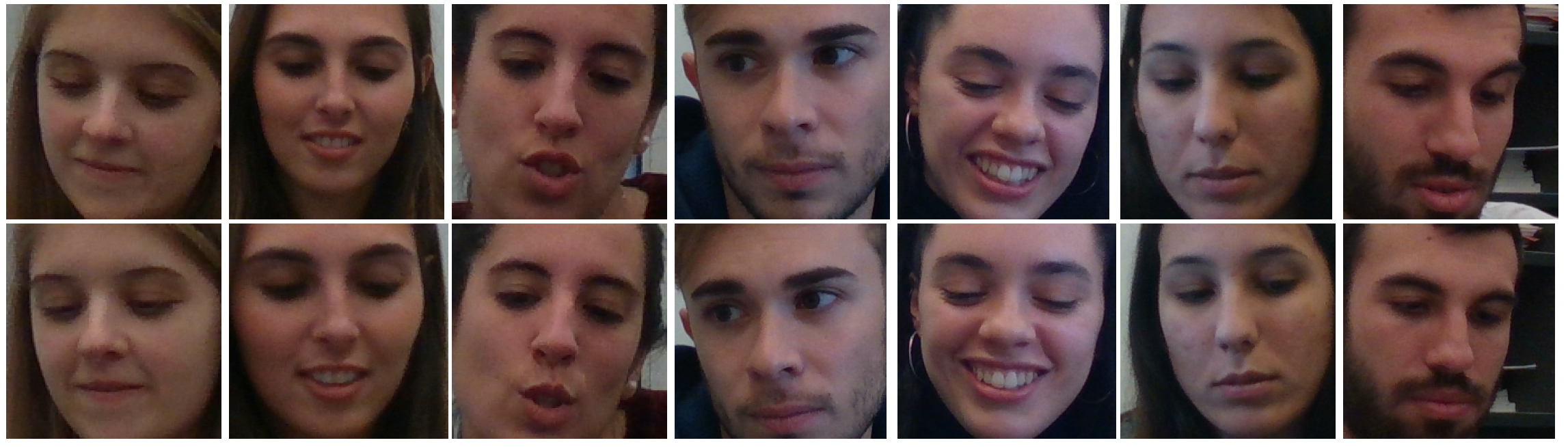}
    \caption{The example of pairs of recording facial area of several participants when telling True (top) and Lies (bottom).}
    \label{fig:faces}
    \end{center}
\end{figure}

The examples of the recorded participants can be seen in the Figure \ref{fig:faces}. Notice that in several examples, the overall facial expressions are relatively similar, so that it could be a challenging tasks for any visual based lie detection algorithm. Thus, using this data, we can expect to perform appropriate test for the effectiveness of current machine learning based lie detection approaches, that we will detail in the next sections. 




\section{Methodology}
\label{sec:method}

We use our recorded GLD datasets to evaluate both classical machine learning approaches and deep learning techniques for this specific lie detection task. In this context, we use facial area as main modality, with the lie labels obtained following our protocol explained on our dataset collections. 


\subsection{Classical Machine Learning}
We use three different handcrafted features that are extracted from RGB facial images: Local Binary Patterns (LBP), Histogram of Oriented Gradients (HOG) and Scale-Invariant Feature Transform (SIFT).
\begin{enumerate}
    \item We follow \cite{lbp_faces} approach to use LBP to model the facial area. Specifically, each image is divided into blocks where LBP is independently computed and resulted in a LBP histogram. The histograms from all images' blocks are then concatenated to form the final descriptor. We also vary the number of (P) use to compute the final descriptor along with the number of blocks to see their impacts. We use the parameter values of P of 8, 12 and 16, with the the number of blocks of 1x1, 2x2, 4x4 and 8x8. For technical implementation, we use Scikit-image library (\cite{scikit-image}) to extract LBP descriptors given the image input. 
    
    \item Given the overall facial area, we calculate the HOG histogram using fixed number of 8 bins, with different number of cells and blocks to test how their impact affects classification performance. We use 1x1 and 2x2 HOG cell sizes, and partition the images into 8x8 and 16x16 pixel sizes. We also use Scikit-image (\cite{scikit-image}) for technical implementation. 
    
    \item We calculated the SIFT features by performing the centroids k clustering that involves the computation of K-Means for all SIFT keypoints. Subsequently, we create the a histogram with K bins for each key-point in an image~(\cite{sift_svm}). In order to evaluate how histogram length and BoW size impacts the classification, we tested different K values: 100, 300, 500 and 800. We used \cite{opencv_library} and \cite{scikit-learn} for the technical implementation. 
    
\end{enumerate}

\begin{figure}[t]
    \centering
    \includegraphics[width=\columnwidth]{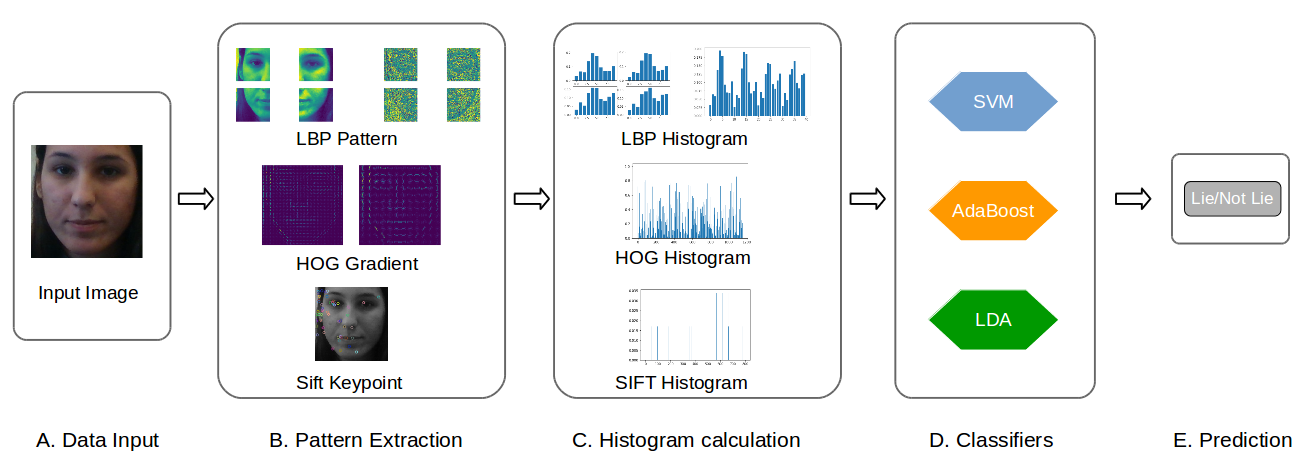}
    \caption{The examples of the pipelines of the Classical Machine Learning for Lie Detection Task. It starts with the input image (a), then the LBP, HOG and SIFT pattern are calculated (b). Subsequently, the histograms of associated pattern are calculated (c) to be used for several classifiers (d) for lie detection (e). }
    \label{fig:classic-archi}
\end{figure}

Using these handcrafted features, then we will employ three classifiers to predict the lie (all implemenation is based on Scikit-learn library (\cite{scikit-learn})  : Support Vector Machine (SVM), AdaBoost, and Linear Discriminant Analysis (LDA). These processes are summarised on the Figure~\ref{fig:classic-archi}.

\subsection{Deep Learning}

For deep learning based approach, we perform transfer learning by means of the embedded features  from  VGG-Very-Deep-16 CNN \cite{vggface}. Specifically, we feed the cropped facial to the pre-trained VGG model, and store the embedded features. Using these embedded features, we then trained the similar classifiers as explained in previous sections to get the baseline results. 

\begin{figure}[t]
    \centering
    \includegraphics[width=\columnwidth]{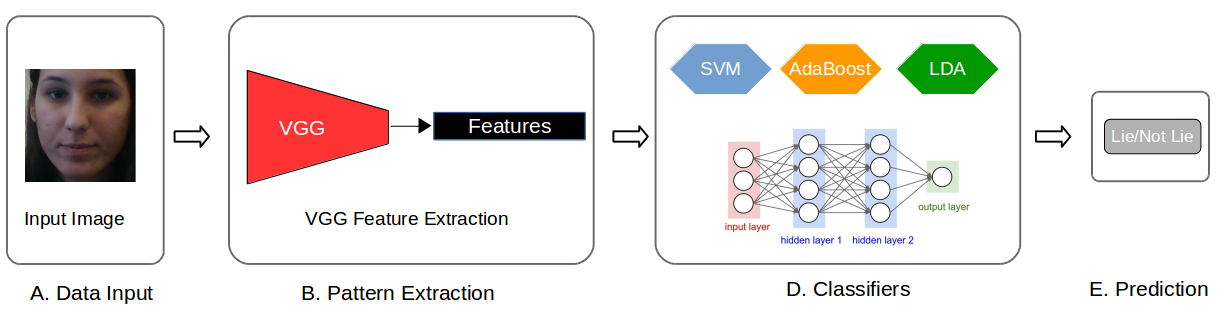}
    \caption{The examples of the deep learning based Lie Detection model. It commences with an input image (a) that is used to calculate the VGG features (b). The VGG features then used by both classical classifiers and Fully Connected Layers (d) for lie detection (e).}
    \label{fig:dl-archi}
\end{figure}

To enable fully trained deep learning model,  we then use the CNN features as an input to the fully connected neural networks consisting two hidden layers with 256 and 128 units respectively and an output layer. Both hidden layers use the rectified linear unit (ReLU) as the activation function, whereas the output unit uses the sigmoid activation function that classifies True (lies) and False (not lies) samples. The model is compiled with Adam optimizer with a learning rate of 0.001 and uses the binary cross-entropy loss. We utilised Keras library \cite{chollet2015} for concrete implementation. Finally, Figure~\ref{fig:dl-archi} shows the overview of these processes.

\subsection{Comparison Metrix}
\label{sub:compMetrix}
We used both Accuracy and F1-Score to judge the quality of the lie estimations of all evaluated approaches. These metrics are calculated as follow: 
\begin{equation}
Accuracy = \frac{TP+TF}{TP+TF+FP+FN}
\end{equation}
\begin{equation}
Precision = \frac{TP}{TP+FP}
\end{equation}
\begin{equation}
Recall = \frac{TP}{TP+FN}
\end{equation}
\begin{equation}
F1 = \frac{2*Precision*Recall}{Precision+Recall} = \frac{2*TP}{2*TP+FP+FN}
\end{equation} 
where TP is True Positive, FP is False Positive, TN is True Negative, and FN is False Negative examples respectively.

\section{Experiments}

We perform three different experiments for the lie detection tasks: Generalization Test, Person Specific Test, and Cross Lie Detection Test. The first experiment evaluates the generalization capacity of the trained lie detector (cf. Section~\ref{sec:method}) to predict the lie status of the never-seen-before participant (i.e not used for training). 

The second test assess the full potential of the lie detector when dealing with a unique participant (i.e customised to a person). This is motivated by the recent report from (\cite{me_not_best_way}) suggesting the personals lying expressions may not be universal. Furthermore, the feel and willingness to perform lying action itself may also differ per-person, that while someone can feel displeased when lying, other people could enjoy it \cite{psychopathy}. Thus by building and testing specialized model for each participant, we can see the theoretical limit of our proposed lie detector. 

Finally, the Real Life Test demonstrate the potential real-life use of the lie detector to deal with different kind of lying conditions and with limited data. This test consists of taking the model with the best performance for both of the previous experiments, and assess their performance with real-time lie detection (from different task). 


\subsection{Generalized models}

\subsubsection{Experiment Settings}
We used our recorded GLD dataset to perform the experiments by splitting the available recording following 5-fold cross validations schemes. We extracted relevant features from both Handcrafted and VGG fatures using the corresponding split. Then used them to train all classifiers (SVM, LDA and FC). Finally, we tested it using the associated test split, and measure the performance using the defined metrics (cf. \ref{sub:compMetrix}). 



\subsubsection{Experiment Results of Classical Machine Learning} 

Table \ref{tab:lbp_results} shows the 5-fold cross validations accuracy and F1 score from LBP descriptors combined with several classifiers (SVM, AdaBoost and LDA). We can see that the best results obtained with the use of Adaboost reaching 52.6\% accuracy and 52 F1 score, which is better than using other classifier such as SVM and LDA. Furthermore, in general notice that the use of 12 points of neighboring (i.e P = 12), and dividing the image with 2x2 grid values produce the best results. This suggests that modest value of parameters are advantageous to improve the lie estimates. 


\begin{table}[t] 
\begin{center}
\caption{Accuracy and F1 for SVM, AdaBoost and LDA for LBP descriptors depending on the number of points (P) and grid size.}
\begin{tabular}{c|cc|cc|cc}
\toprule
\multirow{2}{*}{LBP} & \multicolumn{2}{c|}{SVM} & \multicolumn{2}{c|}{AdaBoost} & \multicolumn{2}{c}{LDA} \\ \cline{2-7} 
 & ACC & F1 & ACC & F1 & ACC & F1 \\ \hline
1x1 Grid – P=8 & 48.8 & 46.4 & 48.8 & 48.6 & 47.8 & 45.2 \\ \hline
2x2 Grid – P=8 & 50 & 46 & 50.2 & 49.6 & 50.2 & 47.4 \\ \hline
4x4 Grid – P=8 & 49.2 & 47 & 49.8 & 48.8 & 49.4 & 47.4 \\ \hline
8x8 Grid – P=8 & 48.4 & 47.2 & 50.6 & 50.4 & 48.6 & 47.4 \\ \hline
1x1 Grid – P=12 & 49 & 47 & 49.2 & 48.8 & 50.4 & 49.6 \\ \hline
2x2 Grid – P=12 & 50.4 & 47.4 & \textbf{52.6} & \textbf{52} & \textbf{51.6} & \textbf{49.8} \\ \hline
4x4 Grid – P=12 & 50.4 & 49 & 49.4 & 48.8 & 50.4 & 49.6 \\ \hline
8x8 Grid – P=12 & 50.4 & \textbf{49.4} & 49.8 & 49.4 & 50.2 & 49.6 \\ \hline
1x1 Grid – P=16 & 49.4 & 47.4 & 50.2 & 49.6 & 49.2 & 48.2 \\ \hline
2x2 Grid – P=16 & \textbf{50.6} & 48.2 & 51.2 & 50.8 & 50.8 & 49.2 \\ \hline
4x4 Grid – P=16 & 50.2 & 46.8 & 48 & 47.4 & 49.6 & 48.4 \\ \hline
8x8 Grid – P=16 & 49.4 & 48.2 & 50 & 49.4 & 49 & 48.4 \\ \midrule
AVG & 49.7 & 47.5 & 49.9 & 49.5 & 49.7 & 48.3 \\ \bottomrule
\end{tabular}%
\label{tab:lbp_results}
\end{center}
\end{table}


We can see the results of HOG descriptor on the Table \ref{tab:hog_results}, that is obtained using similar five-cross validation settings. We can see the similar pattern with the results from LBP, that using 8x8 Grid size with modest value of 2x2 block cells to compute the histogram produce the better results. Furthermore, we note that the best accuracy is achieved by AdaBoost achieving the accuracy 53\% and 52.8 F1 score respectively.

\begin{table}[t]
\begin{center}
\caption{Accuracy and F1 for SVM, AdaBoost and LDA for HOG descriptors depending on cells’ size (8x8, 16x16) and the blocks’ size (1x1, 2x2).}
\begin{tabular}{c|cc|cc|cc}
\toprule
\multirow{2}{*}{HOG} & \multicolumn{2}{c|}{SVM} & \multicolumn{2}{c|}{AdaBoost} & \multicolumn{2}{c}{LDA} \\ \cline{2-7} 
 & ACC & F1 & ACC & F1 & ACC & F1 \\ \hline
8x8 Grid – 1x1 Cells& 50.4 & 50.2 & 50.2 & 50 & 49.6 & 49.4 \\ \hline
8x8 Grid – 2x2 Cells& 51.2 & \textbf{51} & \textbf{53} & \textbf{52.8} & \textbf{51} & \textbf{51} \\ \hline
16x16 Grid – 1x1 Cells& \textbf{52} & 50.4 & 49 & 49 & 48.8 & 48.8 \\ \hline
16x16 Grid – 2x2 Cells & 51 & 49.8 & 51.4 & 51.2 & 48.6 & 48.6 \\ \midrule
AVG & 51.15 & 50.35 & 51.4 & 51.2 & 48.6 & 48.6 \\ \bottomrule
\end{tabular}%
\label{tab:hog_results}
\end{center}
\end{table}

Finally, the results obtained for SIFT descriptors can be seen on the Table \ref{tab:sift_results} with the varying number of the bag of words (BoW ~ K). Here we found that in general, the use of K value of 800 are beneficial. Furthermore, using AdaBoost classifier achieves the maximum results with an accuracy of 53\% and a F1 value of 52.2.

\begin{table}[t]
\begin{center}
\caption{Accuracy and F1 for SVM, AdaBoost and LDA for SIFT descriptors depending on the bag of words size (K).}
\begin{tabular}{c|cc|cc|cc}
\toprule
\multirow{2}{*}{SIFT} & \multicolumn{2}{c|}{SVM} & \multicolumn{2}{c|}{AdaBoost} & \multicolumn{2}{c}{LDA} \\ \cline{2-7} 
 & ACC & F1 & ACC & F1 & ACC & F1 \\ \hline
K=100 & 50 & 49.6 & 50.2 & 49.8 & 50.2 & \textbf{49.6} \\ \hline
K=300 & 49.6 & 49 & 50 & 49.6 & 49 & 48.4 \\ \hline
K=500 & 51.8 & 51 & \textbf{53} & \textbf{52.2} & 50 & \textbf{49.6} \\ \hline
K=800 & \textbf{52.2} & \textbf{51.6} & 52.6 & 51.4 & \textbf{51.6} & 49.4 \\ \midrule
AVG & 51.1 & 50.3 & 51.3 & 50.8 & 49.83 & 49.35 \\ \bottomrule
\end{tabular}%
\label{tab:sift_results}
\end{center}
\end{table}

\begin{figure}[t]
    \centering
    \includegraphics[width=0.45\textwidth]{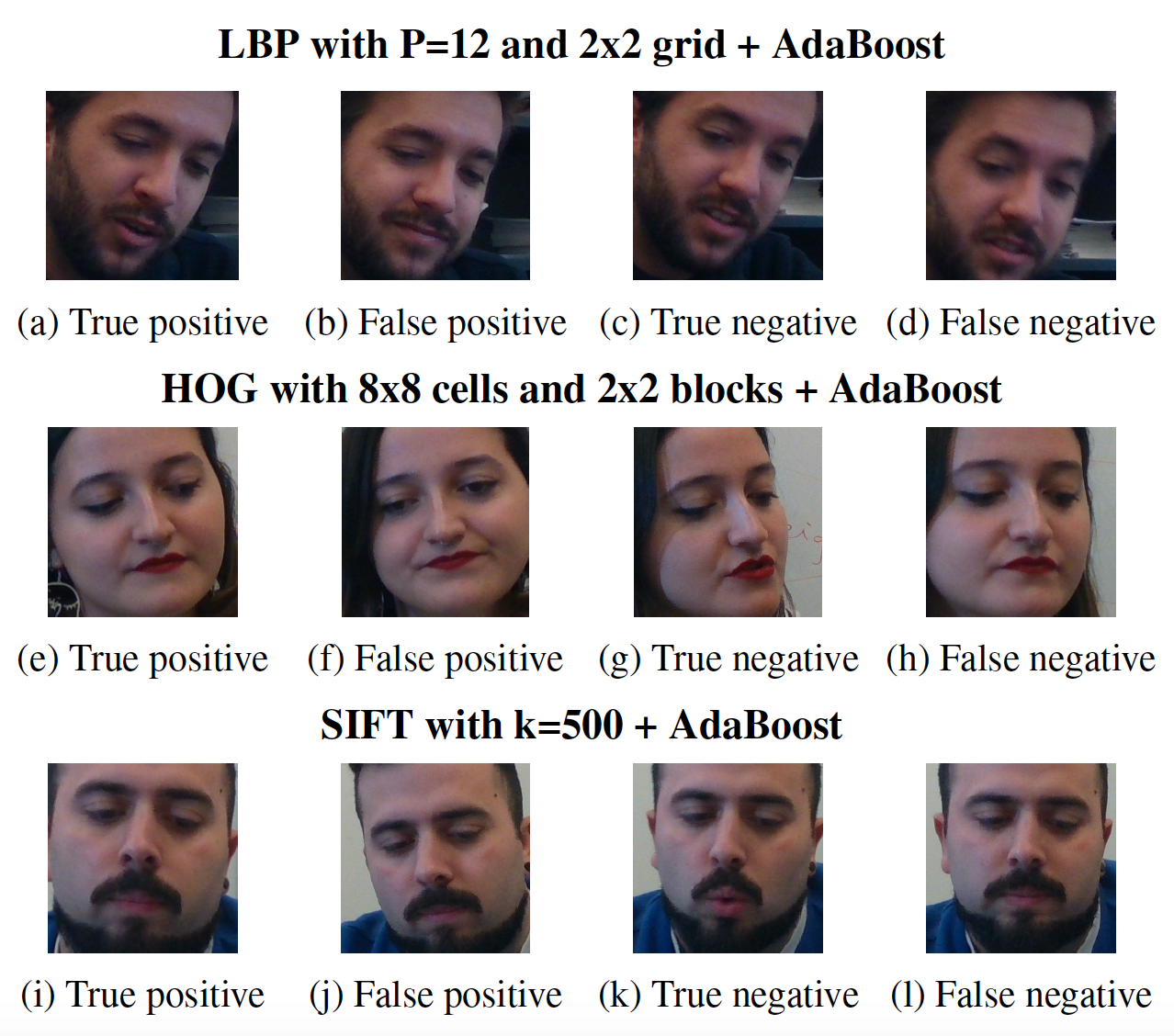}
    \caption{Images examples of correctly and incorrectly classified samples for using handcrafted based features.}
    \label{fig:classical_examples}
\end{figure}

Figure \ref{fig:classical_examples} shows the examples of TP, FP, TN and FN of each best performers of the classical machine learning models.  Notice that the facial expressions are quite similar across the examples, with slight changes happened in the mouth area  in case of both correctly classified lied (TP and FN). Whereas on the failed recogniton (FP, and FN), the facial area are mostly neutrals, thus may confuses the proposed in their predicitons. 


\subsubsection{Experiment Results of Deep Learning} 



\begin{table}[t]
\begin{center}
\caption{Accuracy and F1 achieved using VGG features.}
\begin{tabular}{l|cc}
\toprule
Models & ACC & F1 \\ \hline
VGG+SVM & Na & Na \\ \hline
VGG+LDA & 52 & 50.2 \\ \hline
VGG + AdaBoost & 52.6 & 51.6 \\ \hline
VGG + FC & \textbf{57.4} & \textbf{58.3} \\ \bottomrule
\end{tabular}%
\label{tab:vgg_classical}
\end{center}
\end{table}

\begin{table}[t]
\begin{center}
\caption{Accuracy and F1 achieved with VGG + FC on all five folds.}
\begin{tabular}{ccc}
\toprule
Fold & ACC & F1 \\ \hline
1 & 58 & 56.74 \\ \hline
2 & 56.46 & 48.72 \\ \hline
3 & 54.52 & 58.33 \\ \hline
4 & \textbf{62.76} & 63.79 \\ \hline
5 & 55.49 & \textbf{64.34} \\ \midrule
AVG & 57.44 & 58.38 \\ \bottomrule
\end{tabular}%
\label{tab:vgg_cnn_results}
\end{center}
\end{table}

We present the results of the use of CNN features with both classical classifiers (LDA, Adaboost, SVM) and neural network based classifier of FC on the Table~\ref{tab:vgg_classical}. We can see that results from the use of classical classifiers are quite similar to the results from previous sections, that are modest and suggesting its limitations. Furthermore, we found that using SVM lead to the erroneous value (e.g the lie values are predicted as one class, i.e no change) thus Na value. However, upon the use of FC based classifier, the results are improved reaching 57.4\% and 58.3 F1 value respectively. We need to also note that in one fold, the VGG + FC models were able to reach the 62.76\% and 64.34 F1 value separately, as shown on the Table~\ref{tab:vgg_cnn_results}.  This indicates the compatibility and superiority of Deep Learning based model for this lie detection tasks. 

\begin{figure}[t]
    \centering
    \includegraphics[width=0.45\textwidth]{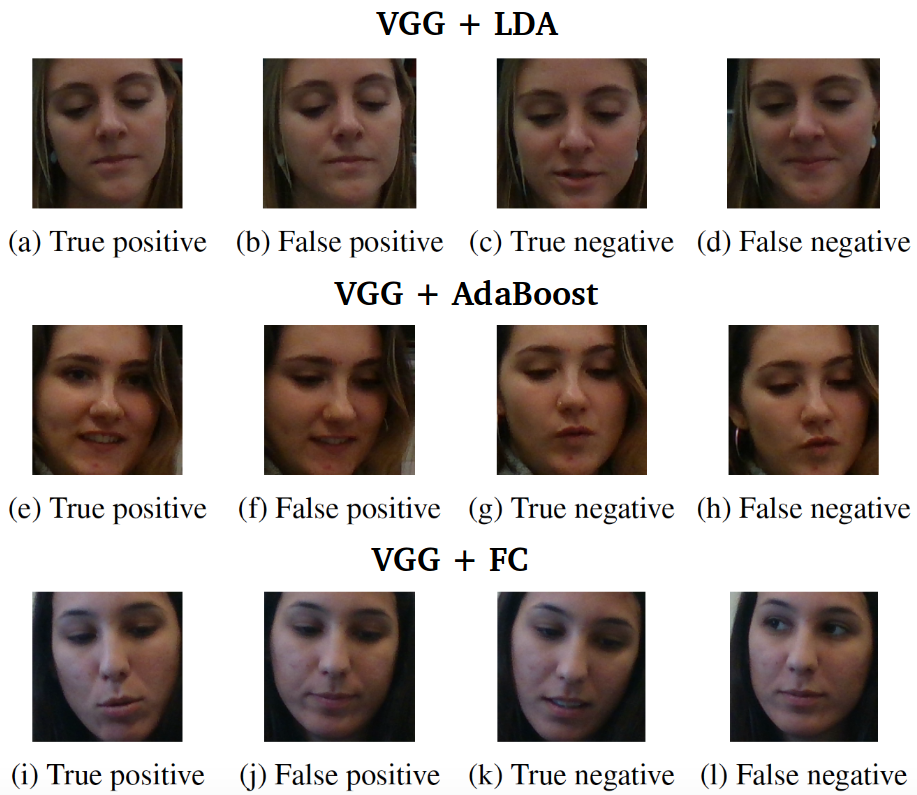}
    \caption{Images examples of correctly and incorrectly classified samples with Deep Learning based Features.}
    \label{fig:deep_results}
\end{figure}

We show on Figure \ref{fig:deep_results} the visual examples of the TP, FP, TN and FN cases of Deep Feature based Lie detector. We can observe that in general, there are more variety in the facial expressions compared to the examples from Classical Machine Learning based detector across examples. We also see that in case failure (FP and FN), the expressions are also more visible compared to Neutral. However, there also seems to be similarity that on the case of the correctly classified label (TP and TN) where the visual changes happen in the mouth area in this example. These variety of expressions suggests the expressiveness of the VGG features which may be helpful to more accurately classify the lie compared to the hand-crafted based descriptor. 





\subsubsection{Overall Comparisons}

\begin{table}[t]
\begin{center}
    
\caption{Highest accuracy and F1 obtained for each descriptor.}
\begin{tabular}{llcc}
\toprule
Methods & Configurations & ACC & F1 \\ \midrule
LBP & P=12, 2x2 grid & 52.6 & 52 \\ \hline
HOG & 8x8 cells, 2x2 blocks & 53 & 52.8 \\ \hline
SIFT + AdaBoost & K=500 & 53 & 52.2 \\ \hline
VGG + FC & 256, 128 and 1 Neuron(s) &\textbf{57.44} & \textbf{58.38}\\
\bottomrule
\end{tabular}%
\label{tab:classical_results}

\end{center}
\end{table}

We can see the overall comparisons of the best performers for all evaluated model in Table \ref{tab:classical_results}. In overall we can see that the results of the classical machine learning technique for lie detection yields quite modest results (close to 50\% accuracy). In other hand, the deep learning based model produces more accurate estimates achieving the best accuracy so far in this dataset of 57.4\% and 58.3 F1 accuracy. Indeed our produced results are quite comparable with the other relevant works for lie detection. Such as the reports from \cite{boxoflies} where the classical machine learning based approach were used (i.e random forest) and  \cite{real-life_trial} where the real-human are employed. 







\subsection{Person specific models}
\label{subsec:person}

\subsubsection{Experiment Settings}
In this experiment, we use the best performer model from previous comparisons (i.e VGG + FC) for individual-based lie detection. We do this by training the model to the equal number of frames to each participants, and tested it on the other frames counterparts. 




\subsubsection{Experiment Results}

\begin{table}[t]
\centering
\caption{Train and test sizes, accuracy and F1 for all participants.}
\resizebox{0.45\textwidth}{!}{\begin{tabular}{l|ccccccccc}
\toprule
\multicolumn{1}{r|}{Metric}    & \multicolumn{1}{c}{p1} & \multicolumn{1}{c}{p2} & p3    & p4    & p5    & p6    & p7    & p8 & p9  
\\\midrule 
ACC & 46.77 & 75 & 45.75 & 79.88 & 87.97 & 56.25 & 62.25 & 74.38 & 24.63\\
F1 & 29.79 & 53.12 & 30.48 & 45.65 & 50 & 9.01 & 38.72 & 51.9 & 26.44
\\\bottomrule
\end{tabular}}
\\[0.3cm]
\resizebox{0.45\textwidth}{!}{\begin{tabular}{l|cccccccc|c}
\toprule
  & p10   & p11   & p12   & p13   & p14   & p15   & p16   & p17 & ALL \\\midrule 
ACC  & 45.38 & \textbf{97.88} & 93.25 & 62.88 & 46.25 & 51.38 & 92 & 46.67 & 65 \\
F1  & 13.01 & \textbf{65.75} & 63.33 & 51.9 & 4.38 & 4.44 & 59.47 & 20.89 & 63.12
\\ 
\bottomrule
\end{tabular}}

\label{tab:individual_results}
\end{table}


Table \ref{tab:individual_results} summarizes the obtained test accuracy for all participants, with column "ALL" contains the mean of the achieved results. Here we can observe that the overall prediction accuracy are higher, with average accuracy of 65\% and F1 score of 63.12 and maximum accuracy of 97.8\% and F1 score of 65.7 in the case of participant 11. This higher accuracy may indicate the ease tasks that the proposed model handles given narrow examples and specialised facial expression that person projects during lying. Thus, further conforms the previously mentioned hypothesis of the unique charactheristics of each persons in performing the lying.





\subsection{Cross Lie Detection Tasks}

\subsubsection{Experiment Settings}
We perform two major cross lie tasks in this experiment that consisted of Card Number Uttering and Sentence Filling. The first test is the simulation of the cards game where the subject holding a deck of cards has to take one card and either utter the real number or to produce the fake number. The second one in other hand, involve the reading of some sentences with blank spaces that have to be filled by the subject with either real or fake information at the time of reading each sentence (the example sentences can be found on the appendix). We perform both of tests by involving a training participant and two test subjects. That is, we first train the model using the data from the training participant when performing both tasks (thus are quite comparable to the person specific task on section \ref{subsec:person}, though now in different task). Subsequently, we used the pre-trained model to detect the lies from the two test subjects when conducting similar tasks. 

\begin{figure}[t]
    \centering
    \includegraphics[width=\columnwidth]{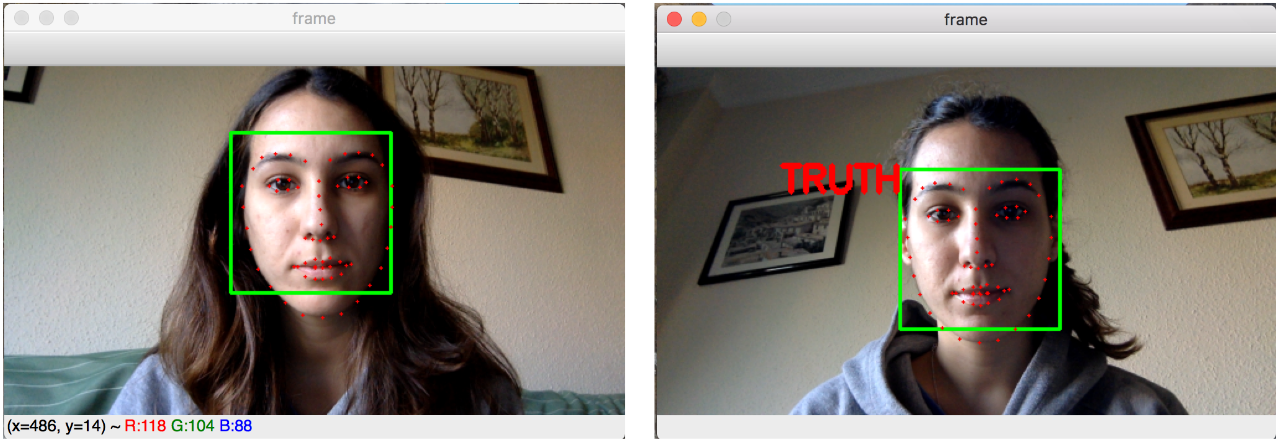}
    \caption{The examples of our proposed program for real-time lie detection.}
    \label{fig:real}
\end{figure}

To collect the samples, we implemented a simple application that integrates different modules: face tracking and cropping (\cite{face_github}), VGG-face 512-dimensional feature prediction  (\cite{keras_vgg_github} and samples prediction as True (lie) or False (not lie). The example of the proposed program can be seen on the Figure~\ref{fig:real}. Using this program on the fly, then we can predict a statement made by the participants. That is, the statement is considered a lie if more than 30\% of the frames are predicted as “lie” by proposed program.

\subsubsection{Experiment Results}


\begin{table}[t]
\begin{center}
\caption{Metric values obtained in real-time evaluation.}
\begin{tabular}{lccc}
\toprule
 & Training & Test Subject 1 & Test Subject 2 \\ \hline
ACC & \textbf{52} & 43.59 & 43.33 \\ \hline
Precision & \textbf{56} & 26.67 & 50 \\ \hline
Recall & 53.85 & \textbf{66.67} & 41.18 \\ \hline
F1 & \textbf{54.9} & 38.1 & 45.16 \\ \bottomrule
\end{tabular}
\label{tab:real-time}
\end{center}
\end{table}



Table \ref{tab:real-time} presents the results obtained from the evaluation for both tasks.  As expected, we can see that the proposed model struggles to correctly predict the true lie label both on the training and test set judged by their low accuracy. Specifically, the best training accuracy of 52\% and F1 score 54.9 of are far lower than person specific test (cf. subsection \ref{subsec:person}) of 65\% and 63.12 respectively. Furthermore, the results of test predictions are also considerable low only reaching 43.59 and F1 score of 38.1. This indicates the difficulty of this prediction task, considering the different charactheristic of the lying condition itself in combinations with personalized way of people during lying. 

\section{Conclusion}

In this paper we presented a comparison of several machine learning based lie detection models applied to our newly collected Game Lie Dataset (GLD). We do so by first collecting the new dataset using several instrumentations and involving 19 participants during the customised card game to incite the lying conditions. Secondly, we pre-processed the data in the structured way to allow for easier loading and future dissemination. Lastly, we cropped the facial area and performed the annotation to complete the dataset productions.

Using our collected dataset, we build classical machine learning models by adopting three handcrafted based features of LBP, HOG and SIFT that later used for lie classification using classical classifier of SVM, Adaboost and LDA. Furthermore, we also include the deep learning based feature of VGG to build a fully end to end system involving fully connected layers to be compared with its semi-classical counterparts by using aforementioned classical classifier for predictions. 

To evaluate the proposed models for lie detection tasks, we performed three main experiments: Generalized Tests, Person Specific Tests, and Cross Lie Detection Tests. On the generalized tests, we found that the limitation of classical methods compared to deep learning based models based on the higher accuracy reached by the latter. Visual inspections also reveal more diverse expression captured by deep learning based model compared to classical approach suggesting its effectiveness. On the second task we show the higher accuracy achieved by our model given its simpler tasks allowing for more accurate learning. That also conforms the hypothesis of the unique facial expression made for each individual during lying. Finally on the last task, we notice the difficulty of the models to properly predict the lie, given the inherent characteristics of the new tasks associated with unique way of lying. 



\section*{Acknowledgement}
This work is partly supported by the Spanish Ministry of Economy and Competitiveness under project grant TIN2017-90124-P, the Maria de Maeztu Units of Excellence Programme (MDM-2015-0502), and the donation bahi2018-19 to the CMTech at UPF.

\appendices
\section{Examples of the Sentences}

1. My name is \underline{\hspace{2cm}}.

2. I was born on \underline{\hspace{0.5cm}}(month), \underline{\hspace{0.5cm}}(day), \underline{\hspace{0.5cm}}(year).

3. I live in  \underline{\hspace{2cm}}.

4. I have \underline{\hspace{2cm}}  siblings.

5. Right now, I am at \underline{\hspace{2cm}}.

6. It is \underline{\hspace{2cm}} (time).

7. My telephone number is \underline{\hspace{2cm}}. 

8. On holidays I am going to \underline{\hspace{2cm}}. 

9. Today I had \underline{\hspace{2cm}} for lunch.

10. Today I woke up at \underline{\hspace{2cm}}.

\ifCLASSOPTIONcaptionsoff
  \newpage
\fi



\bibliographystyle{IEEEtran}
\bibliography{references.bib}
%




%

\begin{IEEEbiography}[{\includegraphics[width=1in,height=1.25in,clip,keepaspectratio]{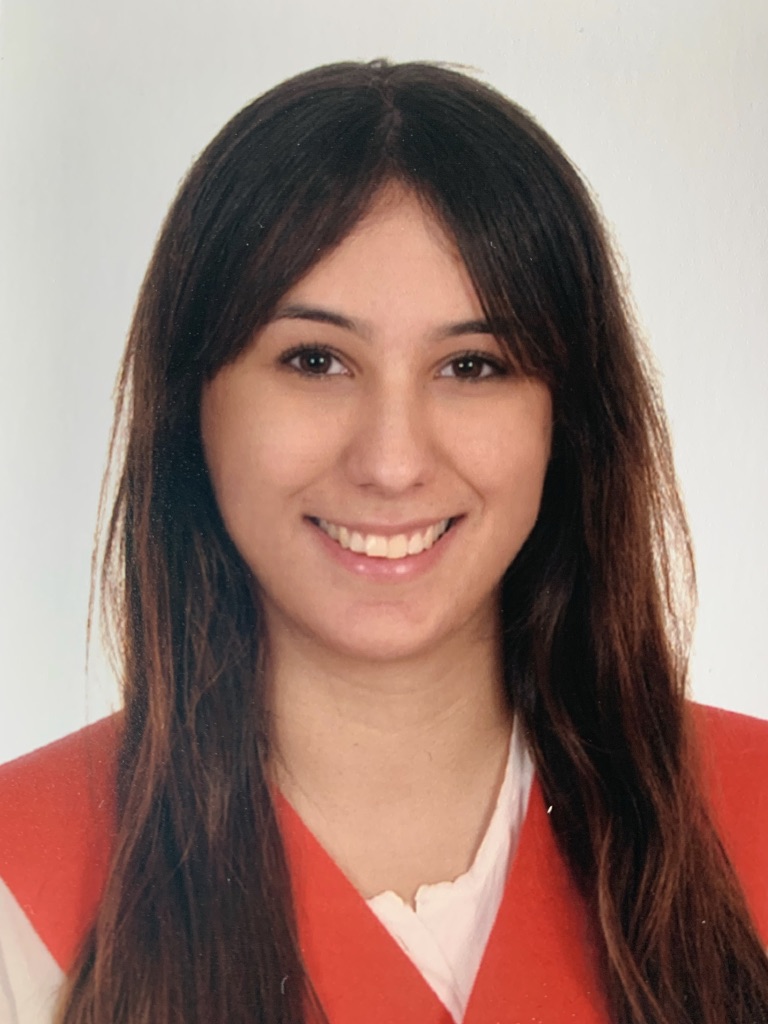}}]{Nuria Rodriguez Diaz} received her Dual Bachelor Degrees in Audiovisual System Engineering and Computer Engineering from Universitat Pompeu Fabra, Barcelona, Spain. She is currently pursuing her Master Degree in MIP Politecnico di Milano, Milan, Italy. 
\end{IEEEbiography}




\begin{IEEEbiography}[{\includegraphics[width=1in,height=1.25in,clip,keepaspectratio]{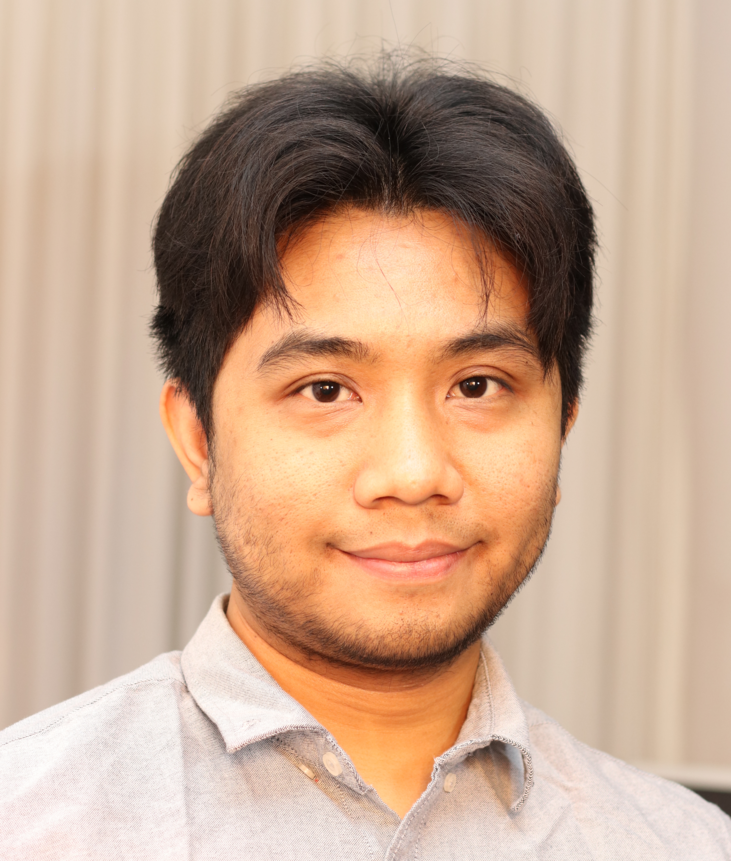}}]{Decky Aspandi} 
received his Bachelor in Computer science from the University of Mulawarman, Indonesia, and M.\,Sc.\ degree in Computer Engineering from King Mongkuts University of Technology Thonburi, Thailand. He received Ph.\,D.\ in Information and Communication Technologies at the Universitat Pompeu Fabra, Barcelona, Spain. He is currently a postdoctoral researcher at University of Stuttgart in Germany, and his research interests include machine learning and deep learning topics with their real life applications.
\end{IEEEbiography}

\begin{IEEEbiography}[{\includegraphics[width=1in,height=1.25in,clip,keepaspectratio]{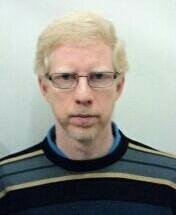}}]{Federico Sukno} received the degree in electrical engineering at La Plata National University (Argentina, 2000) and the Ph.\,D.\ degree in biomedical engineering at Zaragoza University (Spain, 2008). His research activity has been framed in the field of image analysis with statistical models of shape and appearance, targeting diverse applications, most of which in facial biometrics and, to a lesser extent, cardiac imaging. He is currently an Associate Professor in the Department of Information and Communication Technologies at UPF.
\end{IEEEbiography}

\begin{IEEEbiography}[{\includegraphics[width=1in,height=1.25in,clip,keepaspectratio]{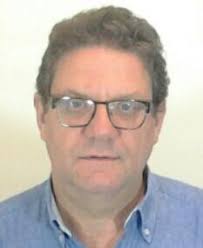}}]{Xavier Binefa} received degrees in Mathematics by Universitat de Barcelona (1976) and in Computer Engineering by Universitat Autònoma de Barcelona (1988). He received his Ph.\,D.\ degree in Computer Vision by Universitat Autònoma de Barcelona in 1996. Currently he is an Associate Professor in the Department of Information and Communication Technologies at Universitat Pompeu Fabra (UPF).
\end{IEEEbiography}




\end{document}